# Automatic Sampling of Geographic objects

Patrick Taillandier[1, 2], Julien Gaffuri[3]

[1]IRD, UMI UMMISCO 209,
32 avenue Henri Varagnat, 93143 Bondy, France
Email: patrick.taillandier@gmail.com

[2]IFI, MSI, UMI 209,
ngo 42 Ta Quang Buu, Ha Noi, Viet Nam

[3] IGN – COGIT laboratory – Paris-Est University
73 avenue de Paris, 94165 Saint-Mandé cedex, France
Email: julien.gaffuri@ign.fr

## 1. Introduction

Many geographical processes require the use of data sample. The method used to choose the geographical object has to be considered. In this paper, we propose approach dedicated to the selection of a sample of geographic objects.

Section 2 presents the context of this work. In section 3, the sampling method we propose is described. Section 4 is dedicated to the evaluation of our method.

## 2  Context

### 2.1  Why using a sampling method?

Many processes are based on the analysis of geographic objects. However, some processes allow to take only into account a limited number of objects. This limitation can come from the implication of an expert in the process (e.g. (Mustière, 2005)) or from the computation complexity of the process (e.g. (Taillandier et al., 2008)). In this context, it becomes important to be able to select a relevant sample of geographic objects among the available ones. Actually, extracting information from an unrepresentative object sample can lead to wrong information or incomplete ones.

### 2.2  Classic sampling methods

The simplest sampling method consists in selecting randomly the needed object number. However, this approach, which is highly stochastic, can lead to unsatisfactory results. Indeed, this approach can lead to select only "similar" objects and miss certain important objects.

Another classic sampling method consists in letting a domain expert selecting a sample of representative objects. This method has two main drawbacks: the first one is it requires the appraisal of an expert; the second one is that the perception of the expert can be biased. Indeed, when thousands of objects can be selected, it is difficult for an expert to have a global view of all the objects.

## 3  Proposed method

### 3.1  Description

The sampling method is composed of three steps detailed henceforward.

*Step 1: Object characterisation*
This step consists in building a *characterised object set* by computing for all objects the values of the measure set. Figure 1 gives an example of such set, on objects of the class "building group".

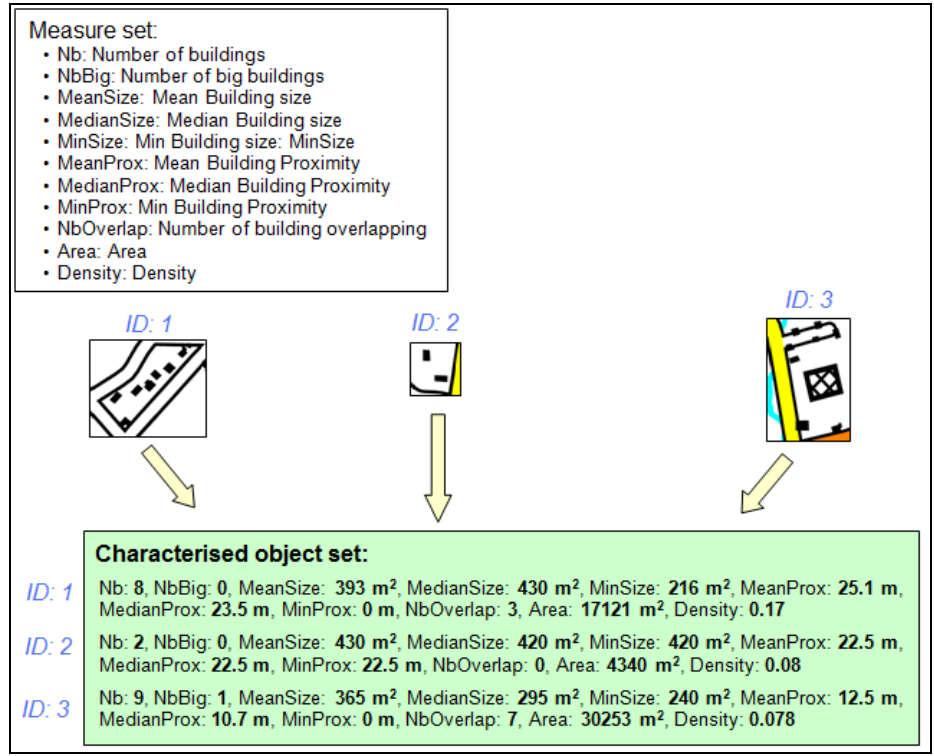

**Fig. 1.** Example of a *characterised object set* for building groups.

*Step 2: Objects clustering*
This step divides the object set into groups of *similar* objects. Defining such groups ensures a representation of all kinds of objects in the object sample.

The object set is divided into groups thanks to clustering techniques used on the *characterised object set*. Most clustering algorithms allow to determine the probability for each object to belong to each group. This probability will be used in the last step.

*Step 3: Object selection*
This last step consists in selecting an objects sample in each group. The selected objects are the most representative of their group, *i.e.* the ones that have the highest probability of belonging to their group.

The number $n_i$ of objects selected in a group *i* depends on the expected size of the sample and on the relative size of the group in terms of the number of objects, compared to the total number of objects. The goal is to keep a proportional representation of the group, and, at the same time, to ensure that even the smallest groups are represented in the sample. The number $n_i$ is computed as follows:

$$n_i = \text{Max}\left(\left\lfloor \frac{\text{sample expected size} \times \text{objects nb in the group i}}{\text{objects total nb}} + 0.5 \right\rfloor, 1\right)$$

### 3.2 Key issues of the method

The method we propose requires different choices that can have a major impact on the sampling result. A first choice concerns the measure set used to characterise the objects. Indeed, as our method bases on this measure set to select the objects, it is important that this one reflects all aspects of the objects and, in the same time do not contains noise or redundant measures. Numerous filtering techniques of irrelevant measures can be used (Mitra et al., 2002). Another important choice concerns the clustering technique and its parameters.

## 4 Sampling method evaluation

### 4.1 General evaluation context

To evaluate the method, we propose an experiment in the context of the automatic revision of procedural knowledge for a cartographic generalisation process.

The generalisation system that we used for the experiment is based on the AGENT model (Barrault et al., 2001). It generalises geographic objects by using a local, step-by-step and knowledge-based method. Each object generalisation is evaluated by a *satisfaction* function whose values are within [1, 10] (10: perfect generalisation). Defining relevant procedural knowledge for this system is a complex task. Thus, in (Taillandier et al., 2008), we proposed an approach to revise it. This approach is based on the analysis of previous experience. It requires an object sample that is used to build experience.

### 4.2 Case study: knowledge revision for building group generalisation

The case study we propose focuses on the generalisation of building groups. The initial data are stemming from BD TOPO®, a 1m resolution database. The target scale is 1:50 000. The initial knowledge base tests all possible actions for each state. Thus, it ensures to find for each generalisation the best possible state considering the constraints and available actions. Nevertheless, it requires exploring many states per generalisation and is thus not efficient enough.

*Sampling method parameters*
Our sampling method requires the definition of different parameters. Concerning the measure set, we defined 11 measures to characterise the building groups (presented Figure 1). For the clustering techniques, we used the well-established EM algorithm (Dempster, 1977).

*Learning and Test areas*
The object sample used by the revision process was selected from a learning area composed of 280 buildings groups. Considering that we chose to use a sample of 50 building groups for the revision process, the number of possible samples is equals to $7 \times 10^{55}$. The quality of the knowledge obtained after revision was evaluated on a test area composed of 200 building groups.

*Object sample defined*
Five object samples were defined: one selected by our method, one selected by a generalisation expert and the last three ones selected randomly. Each sample was composed of 50 building groups selected from the learning area.

*Results*

Figure 2 presents the results after revision with the different knowledge bases. As shown in the figure, the results obtained before revision are good in terms of effectiveness (high satisfaction) but bad in terms of efficiency (too many states tested per generalisation).

After revision with the sample selected by our method, the results obtained are slightly less good in terms of effectiveness, but far better in terms of efficiency. This knowledge base proposes a good balance between effectiveness and efficiency. In comparison, the results obtained with the knowledge base built after revision with the sample selected by the expert are subtly better in terms of effectiveness (but the difference is almost non significant) and worst in terms of efficiency. An explanation is that the expert chose for his sample more complex building groups.

The two first random samples obtained worst results both in terms of effectiveness and efficiency than the sample selected by our method. The last random sample obtained good results in terms of efficiency but bad in terms of effectiveness.

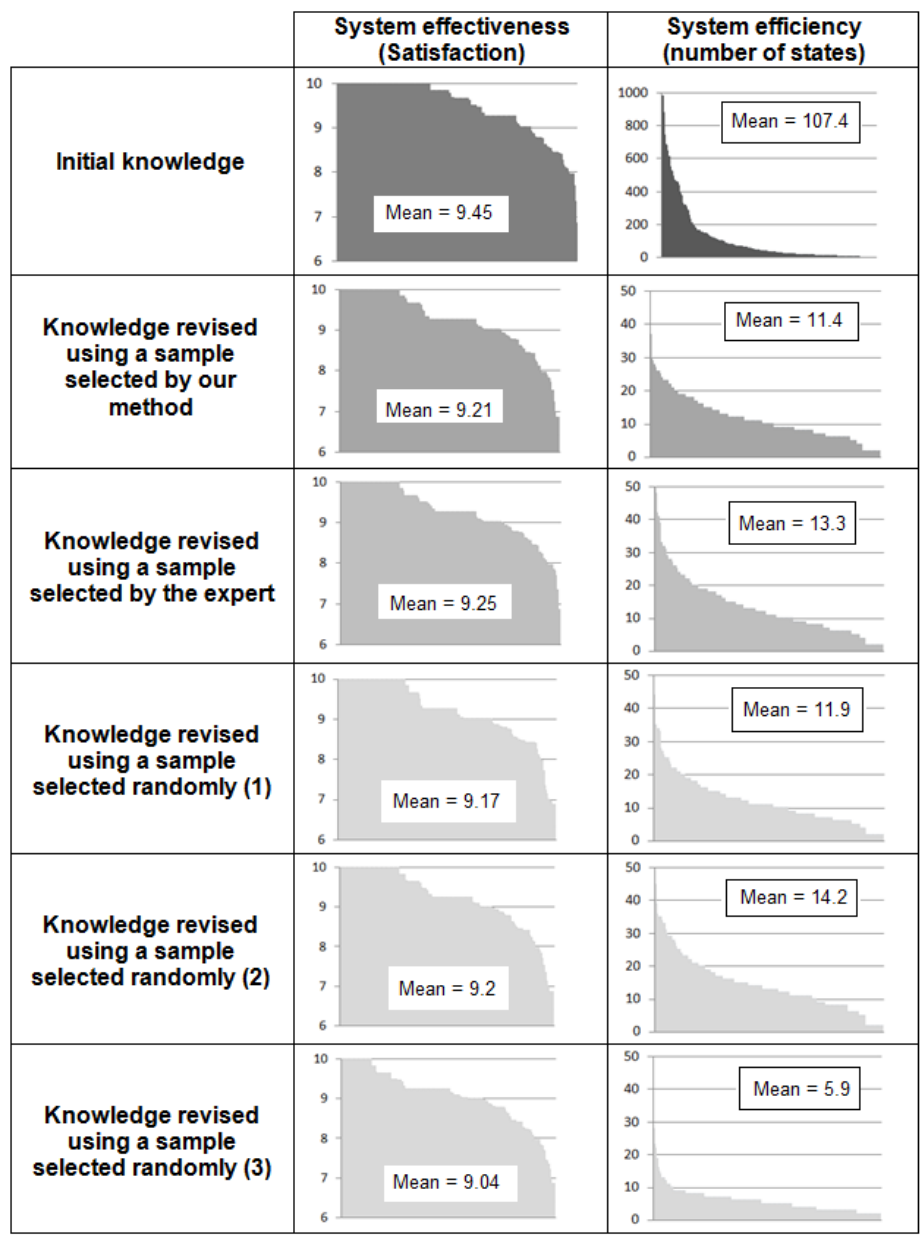

**Fig. 2.** Results obtained by the different knowledge bases

To conclude, our sampling method allowed the best revised knowledge base to be obtained. The sample chosen by the expert allowed to obtain as well a good revised knowledge base. The random sample obtained average or bad results. These results confirm that the choice of the sample is very important for the revision process and show the risk of using a pure random sampling method.

## 5  Conclusion

In this paper, we presented a method dedicated to the sampling of geographic objects. A first experiment, carried out in the context of procedural knowledge revision for a generalisation process, shows that our method is able to select a relevant object sample.

The utilisation of sampling method is a necessity for numerous processes. As our method is generic, an interesting perspective could consist in testing it for other applications.

Some processes require a strong use of domain knowledge. In this context, the intervention of experts can become very important to select a sample of objects. Thus, it could be interesting to adapt our approach in order to better take into account expert knowledge.